\documentclass[final,5p,times,twocolumn]{elsarticle}
\usepackage{xcolor}
\usepackage{algorithm}
\usepackage{algpseudocode}
\usepackage{multirow}
\usepackage{amsmath} 

\usepackage{amssymb}

\begin{document}

\begin{frontmatter}



\title{Basal-Bolus Advisor for Type 1 Diabetes (T1D) Patients Using Multi-Agent Reinforcement Learning (RL) Methodology}


\author[inst1]{Mehrad Jaloli}
\author[inst1]{Marzia Cescon}

\affiliation[inst1]{organization={Department of Mechanical Engineering, University of Houston},
            addressline={4726 Calhoun Rd. }, 
            city={Houston},
            postcode={77004}, 
            state={Texas},
            country={USA}}

\begin{abstract}

This paper introduces a novel multi-agent reinforcement learning (RL) approach for personalized glucose control in individuals with type 1 diabetes (T1D). The proposed methodology utilizes a closed-loop system consisting of a blood glucose (BG) metabolic model and a multi-agent soft actor-critic RL model acting as the basal-bolus advisor. The performance of the RL agents is evaluated and compared to conventional therapy in three different scenarios. The evaluation metrics include minimum, maximum, and mean glucose levels, as well as the percentage of time spent in different BG ranges. Additionally, the average daily bolus and basal insulin dosages are analyzed. The results demonstrate that the RL-based basal-bolus advisor significantly improves glucose control by reducing glycemic variability and increasing the proportion of time spent within the target range of 70-180 mg/dL. Specifically, in scenarios A, B, and C, the time spent within the target range increased from $66.66 \pm 34.97$ $\%$ to $92.55 \pm 4.05$ $\%$, $64.13 \pm 33.84$ $\%$ to $93.91 \pm 6.03$ $\%$, and $58.85 \pm 34.67$ $\%$ to $78.34 \pm 13.28$ $\%$, respectively. The RL-based approach also effectively prevents severe hyperglycemia events (p $\leq 0.05$) and reduces the occurrence of hypoglycemia. For scenarios A and B, hypoglycemic events decreased from $14.2 \% \pm 32.27 \%$ to 3.77$\%$ $\pm$ 4.01$\%$ and $16.59\% \pm 32.42\%$ to 2.63$\%$ $\pm$ 4.09$\%$, respectively. Notably, in scenario C, no hypoglycemic events were experienced in either of the methodologies due to a reduction in insulin sensitivity. Furthermore, the study demonstrates a statistically significant reduction in the average daily basal insulin dosage with the RL agent compared to conventional therapy (p $\leq 0.05$). Overall, these findings indicate the effectiveness of the multi-agent RL approach in achieving better glucose control and mitigating the risk of severe hyperglycemia in individuals with T1D.

\end{abstract}




\end{frontmatter}



\begin{figure*} 
    \centering
    \includegraphics[width=\textwidth]{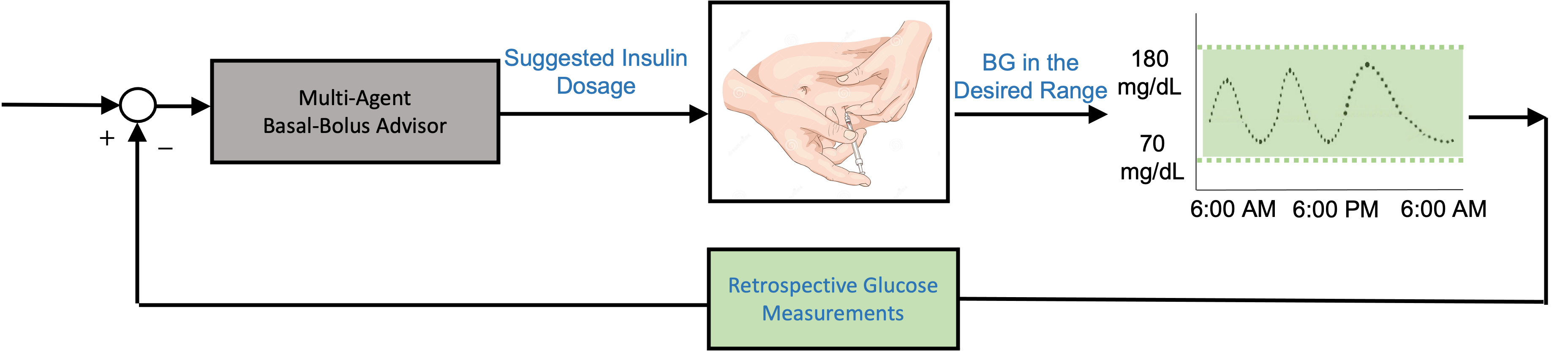}
    \caption{Block Diagram of the proposed closed-loop control of the BG level, using a multi-agent RL-based methodology}
    \label{fig_block_diag}

\end{figure*}
\section{Introduction}
\label{sec:Intro}

Type 1 diabetes (T1D) patients require exogenous insulin administration for survival, but suboptimal glycemic control is a common problem in clinical practice \cite{american202213}. This highlights the urgent need to explore insulin injection protocols that can optimize glycemic outcomes and improve health-related quality of life. Given the substantial inter-patient variability in T1D, it is essential to develop personalized insulin administration techniques \cite{solomon2018sources}.
The basal-bolus method is a widely used approach for managing blood glucose (BG) levels within the euglycemic range. This method involves using basal insulin between meals or at night to regulate fasting BG levels, while bolus doses are administered to compensate for the rise in BG levels after meals  \cite{boiroux2017adaptive}.

In general, the determination of basal doses in diabetes management is often based on a heuristic methodology, incorporating the patient's medical data and the physician's clinical expertise. In contrast, the calculation of meal boluses follows a specific scientific procedure:

\begin{equation}
    u_{bolus} = \frac{CHO}{CR} + \frac{(G_c - G_d)}{CF} - u_{IOB}
\end{equation}
where $CHO$ represents the estimated carbohydrate content of the meal, \(G_c\) and \(G_d\) denote the actual and target blood glucose levels respectively, \(u_{IOB}\) represents the remaining insulin on board (IOB), and \(CR\) and \(CF\) represent personalized insulin-to-carbohydrate ratio and correction factor respectively.

The basal-bolus method may be administered via continuous subcutaneous insulin infusions (CSII) with an insulin pump \cite{linkeschova2002less} or multiple daily injections (MDIs) \cite{cescon2020using}.
Automated insulin delivery (AID) systems, such as the artificial pancreas (AP), are an alternative strategy that combines continuous glucose monitoring (CGM) and insulin pumps with a control algorithm  \cite{breton2012fully,clarke2009closed}. 

While AP systems have demonstrated promising results in glucose management, it is important to note that some individuals undergoing insulin treatment prefer MDI over insulin pumps and AP systems due to the associated burdens and difficulties \cite{tanenbaum2017diabetes}.

On the other hand, optimizing an AID system to implement MDI therapy in individuals with type 1 diabetes (T1D) poses a complex challenge  \cite{thomas2021optimizing}. This involves careful consideration of both basal and bolus insulin doses based on the patient's BG readings. CGM sensors play a crucial role in obtaining frequent BG readings, typically at a sampling rate ranging from 5 to 15 minutes. Additionally, incorporating personalized insulin-to-carbohydrate ratios and correction factors is essential to enhance the accuracy of calculated meal boluses. 

In the field of diabetes management, extensive research has been conducted to develop insulin delivery algorithms for automating the MDI regimen. Various methodologies have been explored in the literature, including PID and fuzzy logic approaches \cite{campos2005self,campos2006fuzzy}, optimization-based methods  \cite{kirchsteiger2009robustness,cescon2012impulsive,carrasco2017design}, and iterative learning strategies \cite{cescon2020using,owens2006run}.

Cescon et al. (2019) explored the application of ILC with sparse measurements for long-acting insulin injections in T1D patients \cite{cescon2019iterative}. They utilized ILC to administer basal insulin through once-a-day dosing of long-acting insulin analogs, incorporating a modified metabolic model that considers subcutaneous insulin kinetics. Simulation results showcased the advantages of this approach, including robust performance in the presence of induced insulin resistance.

Moreover, advances in the development and increase in the popularity of CGM sensors have facilitated the comprehensive collection of multivariable personal health data from various diabetes-related sensors. Consequently, this technological advancement has paved the way for the development of data-driven and artificial intelligence (AI)-based approaches to diabetes management \cite{cescon2023system}. Specifically, machine learning (ML) and neural networks (NN), as subfields of AI, have exhibited significant capabilities in detecting patterns from datasets, enabling accurate predictions \cite{jaloli2022long} and informed decision-making \cite{tyler2020artificial}. These methodologies have been successfully applied in providing insulin dosage recommendations and predicting future blood glucose (BG) levels. Furthermore, patient-related physiological information, including heart rate (HR) and electrodermal activity (EDA), is incorporated to account for factors such as stress and physical activity that can influence BG variation \cite{jaloli2022incorporating}.

However, it is crucial to address the quality of recorded data in data-driven techniques, as datasets collected under real-world conditions often contain errors and missing data points, which can adversely affect the performance of models. To develop a generalized data-driven model for optimal insulin dosing that can accommodate various conditions and patients, carefully designed experiments are required to examine BG variation in response to a wide range of disturbances. Nevertheless, conducting such experiments on human subjects may pose significant risks to their health \cite{griesdale2009intensive}. To overcome these limitations, in-silico simulators have been employed to create a generalized insulin delivery model, allowing for safer and controlled experimentation.

Reinforcement learning (RL) has emerged as a promising approach in healthcare, particularly in critical decision-making tasks such as treatment recommendations and automated insulin dosage systems  \cite{yu2021reinforcement}. RL provides a self-learning framework where an agent interacts with an environment, receives feedback in the form of rewards or penalties, and learns to make optimal decisions. However, the application of RL in clinical scenarios is currently constrained by the challenges of an extended trial-and-error process and the effective handling of constraints. To overcome these limitations, in-silico simulations are employed as valuable tools for the development and validation of RL systems prior to their implementation in clinical studies involving human subjects, enabling refinement and optimization \cite{sutton2018reinforcement}.

In a notable study by Zhu et al. in 2020, a deep RL model was proposed for basal insulin and glucagon release \cite{zhu2020basal}. By continuously optimizing rewards based on glucose outcomes, RL was leveraged to determine insulin dosage in multiple daily injections (MDI) therapy over an extended control period, such as a day or a week \cite{zhu2020insulin}. This approach harnesses the inherent capability of RL to learn optimal control tasks, providing a potential avenue for improving insulin decision-making within MDI therapy.

Furthermore, another key challenges associated with RL lies in effectively handling constraints and mitigating the prolonged trial-and-error training process. Initial actions taken by RL agents may inadvertently trigger complications within the patient's physiological framework, thus restraining the practical applicability of RL in real-world clinical settings. Recent studies have delved into diverse RL methodologies, encompassing actor-critic, Q-learning \cite{noaro2023personalized}, SARSA, and Gaussian process RL \cite{tejedor2020reinforcement}, with a specific focus on glucose control utilizing simulated datasets.

In our previous study \cite{jaloli2023reinforcement}, we introduced an RL model that specifically focused on the precise administration of bolus insulin. Within this model, an RL agent assumed the crucial responsibility of determining the optimal dosage of bolus insulin. In contrast, the delivery of basal insulin was effectively managed by an iterative learning control (ILC) controller, proposed in \cite{cescon2020using}. The integration of this ILC controller within a closed-loop system, along with the RL agent and a blood glucose (BG) simulator, resulted in the formation of a comprehensive closed-loop basal-bolus insulin delivery system tailored for multiple daily injections (MDI) therapy.

Building upon our previous work, the present study proposes an extension by developing a novel multi-agent SAC-RL methodology for basal-bolus dosage administrating for MDI therapy in T1Ds. The block diagram of the proposed methodology is proposed in Fig \ref{fig_block_diag} This advanced methodology is designed to automate the precise administration of basal-bolus insulin by leveraging interactions with the FDA-approved BG simulator, which serves as the simulated environment. The integration of the SAC-RL framework within a closed-loop system enables continuous learning and adaptation, ensuring optimal insulin delivery in a dynamic and personalized manner.

\section{Materials and Methods}
\label{sec: Methods}

This section describes the development of a framework for MDI therapy. It involves integrating a multi-agent RL model into a metabolic model and training it with real-time data. The pretrained multi-agent RL model is then utilized to create an automated closed-loop insulin administration system for MDI therapy. Additionally, this section covers the generation of the training dataset as well as the design of the reward function, for building the RL model.

\subsection{BG Testing Platform}
\label{bg_simulator}

To simulate MDI treatment in individuals with T1D, we utilized a metabolic model of 10 in-silico patients initially introduced by Cescon et al.  \cite{cescon2020using}. It is worth highlighting that the patient's specific parameters were derived from the sources \cite{kovatchev2020method}.
In order to assess the efficacy of the proposed model under varying circumstances, three distinct meal scenarios were investigated. 
\begin{itemize}
    \item \textbf{Scenario A}: This scenario represented the nominal case in which three meals were consumed each day at set times, namely 7 a.m., 1 p.m., and 7 p.m., with a corresponding carbohydrate intake of 50 g, 75 g, and 75 g, respectively. This scenario provided a baseline for the model's performance and served as a reference point for the other scenarios.
    \item \textbf{Scenario B:} In this scenario the model's robustness against meal disturbances was tested, wherein the timing of meals was distributed according to a normal distribution, with a mean of 30 minutes and a standard deviation of 5 minutes. Additionally, the amount of carbohydrate intake for each meal was also normally distributed, with a mean of (50, 75, 75) grams for breakfast, lunch, and dinner, respectively. The standard deviations for carbohydrate intake were (5, 7.5, 7.5) grams for breakfast, lunch, and dinner, respectively. This scenario represented a more realistic situation that one may encounter in real life, where meal timings may be unpredictable, and the amount of carbohydrates consumed at each meal may vary.
    \item \textbf{Scenario C:} This scenario was designed to examine the model's robustness against insulin sensitivity disturbances. In this scenario, meal timings and carbohydrate amounts were maintained from Scenario B while insulin resistance was induced by modifying the parameters in the metabolic model that affect insulin's impact on glucose uptake and endogenous glucose production by 40\%. This scenario allowed for the assessment of the model's ability to perform under conditions of fluctuating insulin sensitivity, which may be a common occurrence in individuals with diabetes.

\end{itemize}

To develop a closed-loop insulin delivery system that accurately represents real-life scenarios and adapts insulin dosages based on real-time glucose readings, we leverage the SAC RL model introduced in \cite{haarnoja2018soft}  and integrate it into the testing platform. By training RL agents using real-time data from the BG simulator across three diverse scenarios, our aim is to enable the agents to dynamically optimize insulin dynamics and enhance the time in range (TIR) within the euglycemic range of 80-170 mg/dL. The utilization of multiple scenarios allows for the generalization and robustness of the trained agents to different real-life situations, ensuring their effectiveness in practical applications.

\subsection{Multi-Agent RL Model }
\label{subsec: rl_model}
 
The multi-agent RL model used in this study consisted of two agents: a basal agent and a bolus agent, each representing different types of insulin. Both agents were trained using the soft actor-critic (SAC) algorithm, which is well-suited for continuous control problems with stochastic dynamics.

The development of the multi-agent RL model followed a two-step process. In the first step, the basal agent was trained in a closed-loop fashion using real-time data from the BG simulator. Specifically, data from scenario A, representing the nominal scenario, was utilized during the 7-hour fasting period before the first meal of the day. During the training of the basal agent, the bolus dosages for each meal were determined using the optimized insulin-to-carbohydrate ratio (ICR) described in the \cite{cescon2020using}. This ICR provides a guideline for calculating the appropriate amount of insulin based on the meal's carbohydrate content. 

This training aimed to evaluate the ability of the prescribed basal dosage to stabilize the BG level after fasting. The training objective involved minimizing an objective function that combines expected cumulative rewards and policy entropy.

Training of the basal agent involves minimizing the following objective function:

\begin{equation}
J(\theta) = \mathbb{E}_{\tau \sim p_{\theta}} \left[ \sum_{t=0}^{T} \gamma^t \left( r_t + \alpha H(\pi_{\theta}(a_t|s_t)) \right) \right],
\end{equation}
where $J(\theta)$ is the objective function to be optimized, $\tau$ is a trajectory of states, actions, and rewards sampled from the environment under policy $\pi_{\theta}$, $p_{\theta}$ is the distribution of trajectories induced by the policy $\pi_{\theta}$, $r_t$ is the reward received at time $t$ obtained from the basal agent's reward function explained in Algorithm 1 in the section \ref{reward_func}. $\gamma$ is the discount factor, $\alpha$ is a hyperparameter that controls the weight of the entropy regularization term, and $H(\pi_{\theta}(a_t|s_t))$ is the entropy of the policy.

In the second step, the pre-trained basal agent was integrated into a model along with the bolus agent and the BG simulator.

The goal was to train the bolus agent using the pre-trained basal agent in the loop such that the action taken by the basal agent could affect the state of the environment, i.e. the BG simulator which consequently required the bolus agent to adapt its action accordingly. The training objective for this step was similar to the previous step, having the effect of both meals and the action taken by the basal agent on the current state of the environment. 

Training of the bolus agent involves minimizing the following objective function:
\begin{equation}
J(\theta) = \mathbb{E}_{\tau \sim p_{\theta}} \left[ \sum_{t=0}^{T} \gamma^t \left( r_t + \alpha H(\pi_{\theta}(a_t,b_t|s_t)) \right) \right],
\end{equation}
where $a_t$ is the action taken by the bolus agents at time $t$, and $\pi_{\theta}(a_t|s_t)$ is the policy of the two agents. $r_t$ is the reward received at time $t$ obtained from the bolus agent's reward function explained in Algorithm 2 in section \ref{reward_func}.

Finally, the performance of the obtained multi-agent basal-bolus RL model was evaluated by exposing it to real-time new sets of data generated by the BG simulator, described in section \ref{sec: results}.

\subsection{Reward Function}
\label{reward_func}
The main goal of this closed-loop model is to increase the time in range (TIR) by reducing the occurrence of hypoglycemia (BG $<$ 70 mg/dL) and hyperglycemia (BG $>$ 180 mg/dL). To achieve this, we have defined two distinct reward functions for each agent.

In the BG simulator, blood glucose (BG) readings are taken at fixed 1-minute intervals. However, to develop a model suitable for MDI therapy, the general idea regarding the timing of the actions taken by each patient, is that the bolus RL agent only takes action when a meal is consumed, while the basal agent takes action once a day at 7:00 AM. We have set the sampling rate to 15 minutes for the bolus agent and 1 day for the basal agent, and have designed separate reward functions for each agent to guide their learning process.

The reward function for the basal agent was designed to evaluate the BG level during the first 7 hours of the day (12:00 AM to 7:00 AM). Three distinct rewards, denoted as $r^{(105, 115)}$, $r^{(100, 120)}$, and $r^{(70, 180)}$, were computed based on the number of occurrences that the measured BG level fell within specific ranges: (105, 115) mg/dL, (100, 120) mg/dL, and (70, 180) mg/dL, respectively. To determine the final reward for each action taken per each episode, which corresponds to a one-week duration, the following equation was employed:

\begin{equation}
    r_{Basal}^{episode} = \sum_{i = 1}^{N} \exp{\left(\frac{r(i)^{(105, 115)}}{2}\right)} \\
    + \exp\left(\frac{r(i)^{(100, 120)}}{2}\right) + \exp\left(\frac{r(i)^{(70, 180)}}{2}\right)
\end{equation}
where, $r_{Basal}^{episode} $ represents the total reward for the action taken by the basal agent on each episode which is obtained by summing all the daily rewards from day 1 to day N in each episode.



This reward function captures the maximum duration during the fasting period in which the BG level remains within the target range. This period is of particular significance as it serves as the baseline for individuals with T1D. The reward is determined based on the agent's actions and reflects its ability to effectively maintain the BG level within the desired range during this fasting period. Algorithm 1 provides the pseudocode for the reward function outlined above.
\begin{algorithm}[!ht]
\caption{Basal Agent Reward Function}
\label{alg:basal_reward_function}
\begin{algorithmic}[1]
\State   Initialize parameters $BG^  {buffer}$, $BG^  {bounds}$
\State   Inputs:
\State $BG^  {buffer}$: A buffer of CGM readings for the past 7 hours.
\State $BG^  {bounds}$: Bounds of the BG readings,i.e. (105, 115) [mg/dL]
\State   Output:
\State $r^  {episode}$: Calculated reward for each episode
\State $nnz \gets$   number of nonzero elementss
\State $n \gets$ number of samples in the $BG^  {buffer} array $
\For{each episode}
	\For{each step, $t=1,\dots,N$}
		\State $r_{t}^{(105, 115)} \gets  \frac{10}{n} \times nnz(BG^  {buffer} < \max(BG^  {bounds}) \land BG^  {buffer}> \min(BG^  {bounds}))$
  
    	\State $r_{t}^{(100, 120)} \gets\frac{10}{n} \times nnz(BG^  {buffer} < (\max(BG^  {bounds})+5) \land BG^  {buffer} > (\min(BG^  {bounds})-5))$

     	\State $r_{t}^{(70, 180)} \gets\frac{10}{n} \times nnz(BG^  {buffer} < (\max(BG^  {bounds})+65) \land BG^  {buffer} > (\min(BG^  {bounds})-35))$

	\EndFor
	\State $r^  {episode} \gets \sum_{t=1}^{N} \exp\left(\frac{r_{t}^{(105, 115)}}{2}\right) +
\exp\left(\frac{r_{t}^{(100, 120)}}{2}\right) +
\exp\left(\frac{r_{t}^{(70, 180)}}{2}\right) $
\EndFor
\end{algorithmic}
\end{algorithm}

The bolus agent considers several factors, including the patient's previous BG value, meal information generated by the simulator, and a history of actions taken in the last three hours. Its primary objective is to maintain the BG level within the euglycemia range of (70, 180) mg/dL.

The underlying principle for the bolus agent is to receive a positive reward when it takes action immediately after mealtimes and successfully keeps the BG level within the euglycemia range. On the other hand, if the bolus agent takes action at any other time or fails to take action after meal intake, it receives a negative reward as a penalty. These rewards are computed every 15 minutes and accumulated at the end of each episode, which corresponds to a one-week duration.

\begin{algorithm}[!ht]
\caption{Bolus Agent Reward Function}
\label{alg:reward_function}
\begin{algorithmic}[1]
\State  Initialize parameters $BG^{buffer}$, $Meal^{buffer}$, $Action^{buffer}$, $BG^{bounds}$
\State  Inputs:
\State $BG^  {buffer}$: A buffer of BG readings for the past 3 hours.
\State $  {Meal}^  {buffer}$: A buffer of previous meal intakes for the past 3 hours.
\State $  {Action}^  {buffer}$: A buffer of previous bolus insulin doses for the past 3 hours.
\State $BG^  {bounds}$: Bounds of the BG readings, i.e., (70, 180) [mg/dL]
\State  Output:
\State $r^  {episode}$: Calculated reward for each episode
\State $BG^  {Target} \gets 125$
\State $BG_{t-1} \gets BG_1^  {buffer}$
\State $  {Meal}_{t-1} \gets   {Meal}_1^  {buffer}$
\State $  {Action}_{t-1} \gets   {Action}_1^  {buffer}$
\For{each episode}
	\For{each step, $t=1,\dots,N$}
		\If{$  {Action}_{t-1} > 0$ and $  {Meal}_{t-1} > 0$}
			\State $r_t^  {Action} \gets 10$
		\ElsIf{$  {Action}_{t-1} \leq 0$ and $  {Meal}_{t-1} \leq 0$}
			\State $r_t^  {Action} \gets 0$
		\Else
			\State $r_t^  {Action} \gets -2$
		\EndIf
		\If{$BG_t \geq \min(BG^  {bounds})$ and $BG_t \leq \max(BG^  {bounds})$}
			\State $r_t^  {BG} \gets 0.1 \times \exp\left(-\frac{\left|BG_t-BG^  {Target}\right|}{100}\right)$
		\Else
			\State $r_t^  {BG} \gets -0.01 \times \left|BG_t-BG^  {Target}\right|$
		\EndIf
	\EndFor
	\State $r^  {episode}\gets r_t^  {BG} + r_t^  {Action}$
\EndFor
\end{algorithmic}
\end{algorithm}

The total reward for the bolus agent per episode is calculated using the equation:

\begin{equation}
r_{Bolus}^{episode} = \sum_{i=1}^{M} r^{BG}(i) + r^{Action}(i),
\end{equation}
where, $r^{BG}(t)$ represents the rewards associated with the BG level being within the range of (70, 180) mg/dL, indicating euglycemia, and $r^{Action}(t)$ is obtained when the bolus agent takes appropriate action in response to the timing and amount of the meal. The total reward for the bolus agent in each episode is obtained by summing up the rewards for all the actions taken in each episode, denoted as $[1, \ldots, M]$.
Detailed explanations and calculations for these two rewards can be found in Algorithm 2.
By providing the bolus agent with this reward function, we are training it to take actions that lead to maintaining the patient's BG level within the desired range, thereby reducing the risk of hypoglycemia and hyperglycemia. The bolus advisor is an important tool for patients with diabetes as it helps them manage their condition effectively, thereby improving their quality of life.

\section{Results}
\label{sec: results}

The evaluation results for the proposed methodology are presented in this section. Conventional therapy, which involves a once-daily injection of basal insulin at a dosage of 0.4 U/kg of body weight, and predefined insulin-to-carbohydrate ratios for the bolus dosage, as defined in the metabolic model \cite{cescon2020using}, is compared to the performance of the multi-agent basal-bolus advisor. The evaluation is conducted for all three scenarios described in Section \ref{bg_simulator}, and the obtained results are presented in Table 1.

To assess the performance of the proposed methodology, a closed-loop system is implemented containing the BG simulator and the pre-trained multi-agent SAC-RL model acting as the basal-bolus advisor. The test simulation protocol spans a duration of 14 days, starting from Day 0 at 7:00 a.m. The performance of the RL agents is evaluated during the second week, specifically from day 8 to day 14. The initial conditions for the simulations include a glucose level of 125 mg/dL, no insulin present in the plasma or subcutaneous depot, and a reference basal glucose concentration of 125 mg/dL.

Table 1 presents the weekly minimum, maximum, and mean glucose levels in mg/dL. Additionally, the percentage of time during the simulation that the BG value fell within five different ranges was investigated. These ranges include severe hypoglycemia (BG $<$ 54 mg/dL), hypoglycemia (BG $<$ 70 mg/dL), euglycemia (70-180 mg/dL), hyperglycemia (BG $>$ 180 mg/dL), and severe hyperglycemia (BG $>$ 250 mg/dL).

Furthermore, the average daily bolus and basal insulin dosages in U were calculated for each patient. The population mean ± standard deviation was then computed and is presented in Table 1. To assess the statistical significance of the difference between the performance of the RL agents and conventional therapy, paired t-tests were conducted at a significance level of 5\%.

With the basal RL agent, the main objective was to maintain a safe baseline for all the patients in all scenarios to avoid hyper- or hypoglycemia while for the bolus RL agent, the primary objective was to reduce glucose variability and enhance the proportion of time glucose levels maintained within the target range of 70-180 [mg/dL].

\begin{table*} [!ht]
\centering
\caption{Metrics for different scenarios}
\label{tab:metrics}
\begin{tabular}{p{2cm}|cp{3cm}p{3cm}c}

\hline
\textbf{Scenarios} &\textbf{Metrics} & 
\textbf{Conventional} & \textbf{Multi Agent} & \textbf{p-value} \\ 
\centering
 & & \textbf{Therapy} & \textbf{RL} &  \\ 
\hline
\centering
\multirow{10}{*}{\rotatebox[origin=c]{90}{Scenario A}}
&min [mg/dL] & $94.02 \pm 42.06$ & $65.91 \pm 16.37$ & 0.07 \\
&max [mg/dL] & $196.58 \pm 78.87$ & $202.77 \pm 20.29$ & 0.77 \\
&mean [mg/dL] & $134.71 \pm 54.1$ & $120.80 \pm 10.9$ & 0.46 \\
&\% time $<$ 54 [mg/dL] & $9.73\pm 24.78$ & $0.20 \pm 0.32$ & 0.18 \\
&\% time $<$ 70 [mg/dL] & $14.2 \pm 32.27$ & $3.77 \pm 4.01$ & 0.47 \\
&\% time $\in$ (70, 180) [mg/dL] & $66.66 \pm 34.97$ & $92.55 \pm 4.05$ & \textbf{0.04} \\
&\% time $>$ 180 [mg/dL] & $19.14 \pm 30.15$ & $3.68 \pm 2.84$ & 0.12 \\
&\% time $>$ 250 [mg/dL] & $0.01\pm 0.04$ & 0 & 0.34 \\
&Avg Daily Bolus [U] & $18.6\pm 6.18$ & $24.00 \pm 16.46$ & 0.19\\ 
&Avg Daily Basal [U] & $24.26 \pm 4.58$ & $34.56 \pm 11.15$ & \textbf{0.02}\\ 
\hline
\centering
\multirow{10}{*}{\rotatebox[origin=c]{90}{Scenario B}}
&min [mg/dL] & $92.21  \pm 36.9$ & $68.87 \pm 15.26$ & \textbf{0.05} \\
&max [mg/dL] & $201.4 \pm 74.08$ & $201.34 \pm 22.71$ & 0.98 \\
&mean [mg/dL] & $133.84 \pm 50.75$ & $122.33 \pm 6.31$ & 0.53 \\
&\% time $<$ 54 [mg/dL] & $11.23 \pm 24.49$ & $0.29 \pm 1.18$ & 0.25 \\
&\% time $<$ 70 [mg/dL] & $16.59 \pm 32.42$ & $2.63 \pm 4.09$ & 0.28 \\
&\% time $\in$ (70, 180) [mg/dL] & $64.13 \pm 33.84$ & $93.91 \pm 6.03$ & \textbf{0.03} \\
&\% time $>$ 180 [mg/dL] & $19.28 \pm 26.45$ & $3.94 \pm 4.90$ & 0.14 \\
&\% time $>$ 250 [mg/dL] & $3.01 \pm 8.56 $ & $0.06 \pm 0.16$ & 0.35 \\
&Avg Daily Bolus [U] & $18.89 \pm 6.02$ & $24.34 \pm 16.48$ & 0.21 \\
&Avg Daily Basal [U] & $24.00  \pm 4.33$ & $33.17 \pm 11.74$ & \textbf{0.03}\\ 
\hline
\centering
\multirow{10}{*}{\rotatebox[origin=c]{90}{Scenario C}}
&min [mg/dL] & $136.14 \pm 43.74$ & $102.12 \pm 17.92$ & \textbf{0.03} \\
&max [mg/dL] & $243.65 \pm 50.66$ & $224.56 \pm 23.82$ & 0.29 \\
&mean [mg/dL] & $181.54\pm 44.72$ & $157.68 \pm 21.151$ & 0.17 \\
&\% time $<$ 54 [mg/dL] & $0$ & $0$ & $1$ \\
&\% time $<$ 70 [mg/dL] & $0$ & $0$ & $1$ \\
&\% time $\in$ (70, 180) [mg/dL] & $58.85 \pm 34.67$ & $78.34 \pm 13.28$ & \textbf{0.05} \\
&\% time $>$ 180 [mg/dL] & $41.15 \pm 34.66$ & $21.65 \pm 13.28$ & \textbf{0.05 }\\
&\% time $>$ 250 [mg/dL] & $15.65 \pm 32.80$ & $0.08 \pm 0.15$ & 0.15 \\
&Avg Daily Bolus [U] & $18.74 \pm 6.14$ & $27.66 \pm 20.79$ & 0.13 \\
&Avg Daily Basal [U] & $24.26 \pm 4.58$ & $34.36 \pm 7.81$ & \textbf{0.01}\\ 
\hline
\end{tabular}
\end{table*}

\begin{figure*} 
    \centering
    \includegraphics[width=\textwidth]{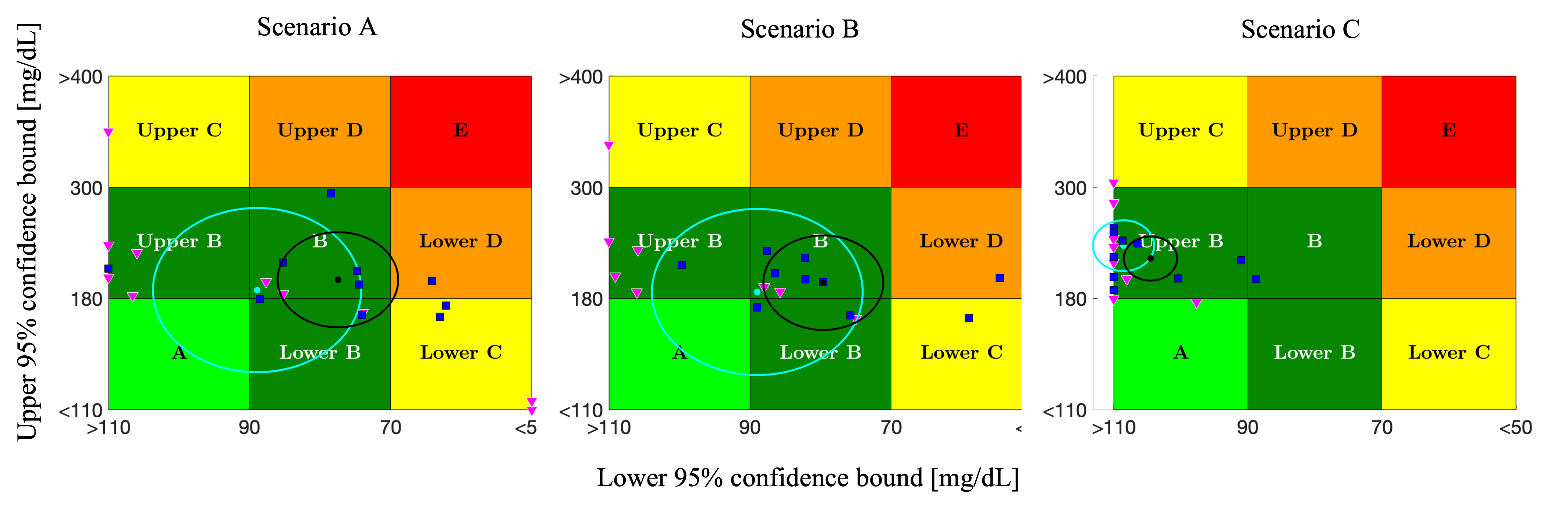}
    \caption{Performance analysis of the three scenarios is conducted based on a blind Continuous Glucose Monitoring (CGM) analysis using CVGA plots. Each marker point on the plot represents the coordinates associated with a single patient. The magenta triangle corresponds to patients undergoing conventional therapy, while the blue square represents those utilizing the proposed RL-based basal-bolus advisor. The dots and circles on the plot indicate the mean and standard deviation, respectively. For the conventional therapy group, the mean and standard deviation are depicted in cyan, while for the RL-based advisor group, they are shown in blue.}
    \label{fig:cvga}
\end{figure*}

Accordingly, Figure \ref{fig:cvga} illustrates the Control-Variability Grid Analysis \cite{magni2008evaluating} that was used to show the comparison between the performances of the proposed strategy against the conventional therapy.


In both Scenario A and B, the multi-agent RL-based basal-bolus advisor demonstrates the ability to concentrate the data points around the middle of the green zone, specifically in the bottom-left area of the grid. Additionally, while there is a slight shift in the mean of the data points compared to conventional therapy, the standard deviation (represented by the radii of the circles) is significantly smaller in the proposed model.

In Scenario C, the performance of the RL-based basal-bolus advisor is particularly notable. The advisor demonstrates superior capability in maintaining both the mean and standard deviation within the green zones, showcasing the robustness of the proposed model against variations in insulin sensitivity compared to conventional therapy.

Furthermore, it is worth mentioning that conventional therapy experiences a few instances of severe hyperglycemic events, indicated by data points located near the boundary between the Upper B and Upper C zones, as well as within the Upper C zone. This further highlights the advantage of the proposed model in effectively preventing severe hyperglycemic events.

It is noticeable that the use of the RL agent resulted in a reduction in glycemic variability across the population such that the population average BG levels reduced from 134.71 $\pm$ 54.1  [mg/dL] to 120.80 $\pm$ 10.9 [mg/dL], 133.84 $\pm$ 50.75 [mg/dL] to 122.33 $\pm$ 6.31 [mg/dL], and 181.54 $\pm$ 44.72 [mg/dL] to 157.68 $\pm$ 21.151 [mg/dL] for scenario A to C, respectively.

In addition, the average percentage of time during the simulation that participants had hyperglycemia (BG $>180$ [mg/dL]) reduced in all three situations, from 19.14 $pm$ 30.15 $\%$ to 3.68 $\pm$ 2.84$\%$ in scenario A, from 19.28 $\pm$ 26.45 $\%$ to 3.94 $\pm$ 4.90 $\%$ in scenario B, and from 41.15 $\pm$ 34.66$\%$ to 21.65 $\pm$ 13.28$\%$ in scenario C. Interestingly, the decreases in the average daily basal units in all scenarios were statistically significant (p = 0.02 and p=0.03 and p=0.01 respectively).

Furthermore, the study demonstrated that the administration of insulin doses by the RL agent led to a statistically significant increase in the proportion of time glucose levels were within the target range for all situations. The percentage of time spent inside the goal range improved from 66.66 $\pm$ 34.97 [$\%$] to 92.55 $\pm$ 4.05 [$\%$] (p=0.04) in scenario A, from 64.13 $\pm$ 33.84 [$\%$] to 93.91 $\pm$ 6.03 [$\%$] (p=0.03) in scenario B, and from 58.85 $\pm$ 34.67 [$\%$] to 78.34 $\pm$ 13.28 [$\%$] (p=0.05) in scenario C.

Overall, as illustrated in both Fig. \ref{fig:cvga} and Table \ref{tab:metrics}, the results obtained from all scenarios indicate the effectiveness of the proposed multi-agent RL-based basal-bolus advisor in achieving better glucose control, reducing glucose variability, and mitigating the risk of severe hyperglycemia.

\section{Discussion and Conclusion}

This study presents a novel closed-loop insulin delivery system for the treatment of patients with type 1 diabetes (T1D) using multiple agent SAC RL-based models integrated into a metabolic model. The system dynamically adjusts insulin dosages based on real-time glucose readings, aiming to optimize glucose control and improve patient outcomes.

The evaluation of the RL agents involved training them using real-time simulated data from 10 in-silico patients with T1D. Subsequently, the agents were tested using real-time data generated by the BG simulator under three different scenarios: nominal, robustness against meal time and carbohydrate amount disturbances, and robustness against insulin sensitivity disturbances.

The results obtained from the simulations highlight the superior performance of the proposed RL agents compared to conventional therapy. The RL agents demonstrate their ability to reduce glucose variability and increase the percentage of time that glucose levels remain within the target range. This improvement in glucose control is a significant step toward managing T1D more effectively.

Furthermore, the proposed model exhibits robustness against meal disturbances and fluctuations in insulin sensitivity, suggesting its potential for real-world applications. The ability to adapt to different scenarios and variations in patient conditions is crucial for achieving optimal glucose control in diverse clinical settings.

While simulation studies provide valuable insights, it is important to acknowledge the need for further research and validation in clinical settings. Conducting clinical trials where T1D patients use the closed-loop system for an extended duration would provide essential data on the system's usability and effectiveness in real-world scenarios. Patient feedback and the assessment of safety and reliability are crucial aspects that need to be evaluated before considering widespread implementation.

To address safety concerns, it is imperative to incorporate fail-safe mechanisms and real-time monitoring into the system. Alarms and alerts can notify healthcare professionals or patients of potential safety issues, allowing for timely intervention. Manual overrides should be available in critical situations, ensuring that healthcare professionals can intervene if necessary. Continuous monitoring of the system's performance and regular checks for abnormal behavior or anomalies will help identify any issues and ensure patient safety.

To enhance the practicality of the proposed model for clinical use, user-friendly interfaces and clear visualizations of glucose data and insulin dosing recommendations should be developed. Providing comprehensive information about the system's operation and decision-making process will facilitate understanding and trust among healthcare professionals and patients. This will foster greater acceptance and adoption of the system in clinical practice.

In conclusion, the proposed closed-loop insulin delivery system based on multiple agent SAC RL-based models shows significant potential in improving glucose control for patients with T1D. However, further research, including clinical trials, validation studies, and safety assessments, is necessary to validate the system's effectiveness and ensure its suitability for real-world clinical use. By addressing safety concerns, optimizing practicality, and conducting rigorous research, the proposed model has the potential to revolutionize glucose control and alleviate the burden of diabetes management for patients undergoing multiple daily injections therapy.

\section*{Acknowledgments}
This work was supported by the University of Houston-National Research University Fund (NRUF): R0504053.

 \bibliographystyle{elsarticle-num} 
 \bibliography{manuscript_new}





\end{document}